\documentclass[conference]{IEEEtran}
\IEEEoverridecommandlockouts
\usepackage{cite}
\usepackage{amsmath,amssymb,amsfonts}
\usepackage{algorithmic}
\usepackage{graphicx}
\usepackage{textcomp}
\usepackage{todonotes}
\usepackage{xcolor}
\usepackage{subfig}
\def\BibTeX{{\rm B\kern-.05em{\sc i\kern-.025em b}\kern-.08em
    T\kern-.1667em\lower.7ex\hbox{E}\kern-.125emX}}

\title{\LARGE \bf
Pic2Diagnosis: A Method for Diagnosis of Cardiovascular Diseases from the Printed ECG Pictures
}

\author{Oğuzhan Büyüksolak$^{1}$ and İlkay Öksüz$^{2}$
\thanks{*This study was supported by the Health Institutes of Turkiye (TUSEB) 2022-EKG-01 Program (Project No. : 20101).}
\thanks{$^{1}$Oğuzhan Büyüksolak is with Computer Engineering Department,
        Istanbul Technical University, Istanbul, Turkey
        {\tt\small buyuksolak20@itu.edu.tr}}%
\thanks{$^{2}$İlkay Öksüz is with Computer Engineering Department,
        Istanbul Technical University, Istanbul, Turkey
        {\tt\small oksuzilkay@itu.edu.tr}}%
}

\begin{document}
\maketitle
\begingroup
\renewcommand\thefootnote{}\footnote{
\textcopyright~2025 IEEE. Personal use of this material is permitted. 
Permission from IEEE must be obtained for all other uses, in any current or future media, 
including reprinting/republishing this material for advertising or promotional purposes, 
creating new collective works, for resale or redistribution to servers or lists, or reuse of any copyrighted component of this work in other works.

To appear in: Proceedings of the 47\textsuperscript{th} Annual International Conference of the IEEE Engineering in Medicine and Biology Society (EMBC), 2025.
}\addtocounter{footnote}{-1}
\endgroup

\thispagestyle{empty}
\pagestyle{empty}

\begin{abstract}
The electrocardiogram (ECG) is a vital tool for diagnosing heart diseases. However, many disease patterns are derived from outdated datasets and traditional stepwise algorithms with limited accuracy. This study presents a method for direct cardiovascular disease (CVD) diagnosis from ECG images, eliminating the need for digitization. The proposed approach utilizes a two-step curriculum learning framework, beginning with the pre-training of a classification model on segmentation masks, followed by fine-tuning on grayscale, inverted ECG images. Robustness is further enhanced through an ensemble of three models with averaged outputs, achieving an AUC of 0.9534 and an F1 score of 0.7801 on the BHF ECG Challenge dataset, outperforming individual models. By effectively handling real-world artifacts and simplifying the diagnostic process, this method offers a reliable solution for automated CVD diagnosis, particularly in resource-limited settings where printed or scanned ECG images are commonly used. Such an automated procedure enables rapid and accurate diagnosis, which is critical for timely intervention in CVD cases that often demand urgent care.
\end{abstract}

\section{Introduction}
Cardiovascular diseases (CVDs) remain the leading global cause of death, highlighting the critical need for accessible and accurate diagnostics. While ECGs are vital for diagnosing CVDs, current machine learning approaches often depend on high-quality digitized signals or artifact-free images, which limits their applicability in real-world scenarios where images may include artifacts like blurring, rotations, or contrast variations.

In many clinical environments, especially in resource-constrained regions, access to advanced ECG devices capable of directly producing digitized waveform data is limited. Instead, ECG outputs are often printed or scanned for analysis and storage. This practice arises from using older ECG machines, the need for physical records, and the lack of infrastructure to manage and store digital signals. As a result, a significant portion of available ECG data exists in non-digital formats, which are prone to artifacts introduced during printing, scanning, or manual handling.

In this work, we present a novel method for directly diagnosing CVDs from ECG images, bypassing the need for digitization. Our deep learning model is trained on diverse datasets containing real-world artifacts, ensuring robustness in practical and resource-constrained settings. By eliminating the digitization step, our approach simplifies diagnostics while maintaining high accuracy, expanding access to reliable cardiovascular care across diverse clinical environments.

\section{Related Work}
\begin{figure*}[htbp]
\centerline{\includegraphics[width=\textwidth]{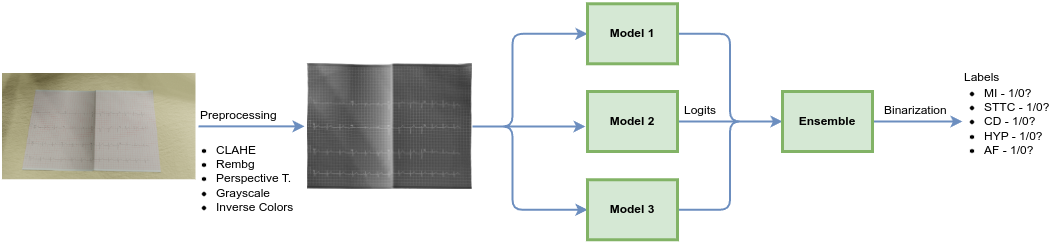}}
\caption{This figure outlines the proposed framework for multi-label cardiovascular disease diagnosis from ECG images. The pipeline begins with preprocessing steps, including CLAHE for contrast enhancement, background removal, perspective transformation, grayscale conversion, and color inversion. The processed image is passed through three independently trained models, and their logits are combined using an ensemble approach. Finally, binarization with per-class thresholds is applied to produce a 1/0 output for each class, indicating the presence or absence of myocardial infarction (MI), ST-T changes (STTC), conduction disorders (CD), hypertrophy (HYP), and atrial fibrillation (AF).}
\label{fig}
\end{figure*}

Cardiovascular disease diagnosis has significantly benefited from advancements in electrocardiogram (ECG) analysis, transitioning from traditional methods to digital and machine-learning-based approaches. Historically, printed ECGs were the standard, leading to the development of reliable digitization methods to integrate these datasets into modern analytical pipelines.

Fortune et al.\cite{b3} introduced an open-source tool for paper-to-digital ECG conversion, addressing challenges such as asynchronous ECG leads in clinical datasets. Their method demonstrated high correlation between digitized and original signals, providing a robust foundation for subsequent analyses. Similarly, Baydoun et al.\cite{b4} developed a MATLAB-based tool for high-precision digitization, enabling legacy paper ECGs to be incorporated into machine learning workflows with over 95\% precision.

Deep learning has further advanced ECG analysis. Wu et al.\cite{b5} proposed a fully automated digitization algorithm that eliminated the need for manual intervention, achieving up to 99\% correlation with original digital signals. Their method allowed for large-scale digitization of historical datasets, facilitating the use of machine learning on previously inaccessible data. Ao and He \cite{b6} extended this by developing deep learning models for direct diagnosis from ECG images, bypassing the need for raw signal processing. However, their model was trained on clean, artifact-free ECG images, limiting its robustness in real-world scenarios.

Our approach diverges significantly by enabling direct diagnosis of cardiovascular conditions without requiring digitization or preprocessing of ECG data. Unlike previous works, which rely on clean ECG images, our method was trained on datasets containing diverse artifacts such as blurring, rotational distortions, and contrast variations. This ensures robustness in clinical environments where imperfections in ECG images are inevitable, especially in resource-constrained or emergency settings. By forgoing the need for digitization and incorporating artifacts in the training process, our model simplifies the diagnostic workflow while maintaining high diagnostic accuracy.

\section{Materials}
This study utilizes two datasets, the first being the well-known PTB-XL dataset~\cite{b2}. PTB-XL contains 10-second recordings of 12-lead ECGs, accompanied by diagnostic annotations provided by two cardiologists. It consists of 21,799 samples collected from 18,869 patients. The second dataset is GenECG~\cite{b7}, a synthetically generated ECG image dataset created using the ECG data from PTB-XL. GenECG has been validated through a clinical Turing test, demonstrating its realistic quality. Since it is derived from PTB-XL, the GenECG dataset also comprises 21,799 samples. It was made available by the BHF Data Science Centre ECG Challenge~\cite{b11}, where the challenge version included five binary labels indicating the presence of specific cardiovascular disease (CVD) classes: MI (Myocardial Infarction), STTC (ST/T changes), CD (Conduction Delay), HYP (Hypertrophy), and AF (Atrial Fibrillation). A total of 15,010 labeled samples were provided, while an additional 3,219 unlabeled samples were designated as the hold-out test set.

In this study, the labeled samples were split into training and validation sets, with 90\% used for training and 10\% for validation.

\section{Methods}

A summary of the proposed method is presented in Fig.~\ref{fig}. The method can be outlined as follows: first, raw ECG images are preprocessed to be used by the classifier models. Subsequently, three classifier models with the same architecture individually produce logits from the preprocessed grayscale images. These logits are then averaged and binarized using predetermined per-class thresholds. The final output is a binary label for each class, indicating the presence or absence of the corresponding cardiovascular disease. The following subsections provide a detailed explanation of the proposed method.

\subsection{Preprocessing}
\begin{figure}[htbp]
\centerline{\includegraphics[width=8cm]{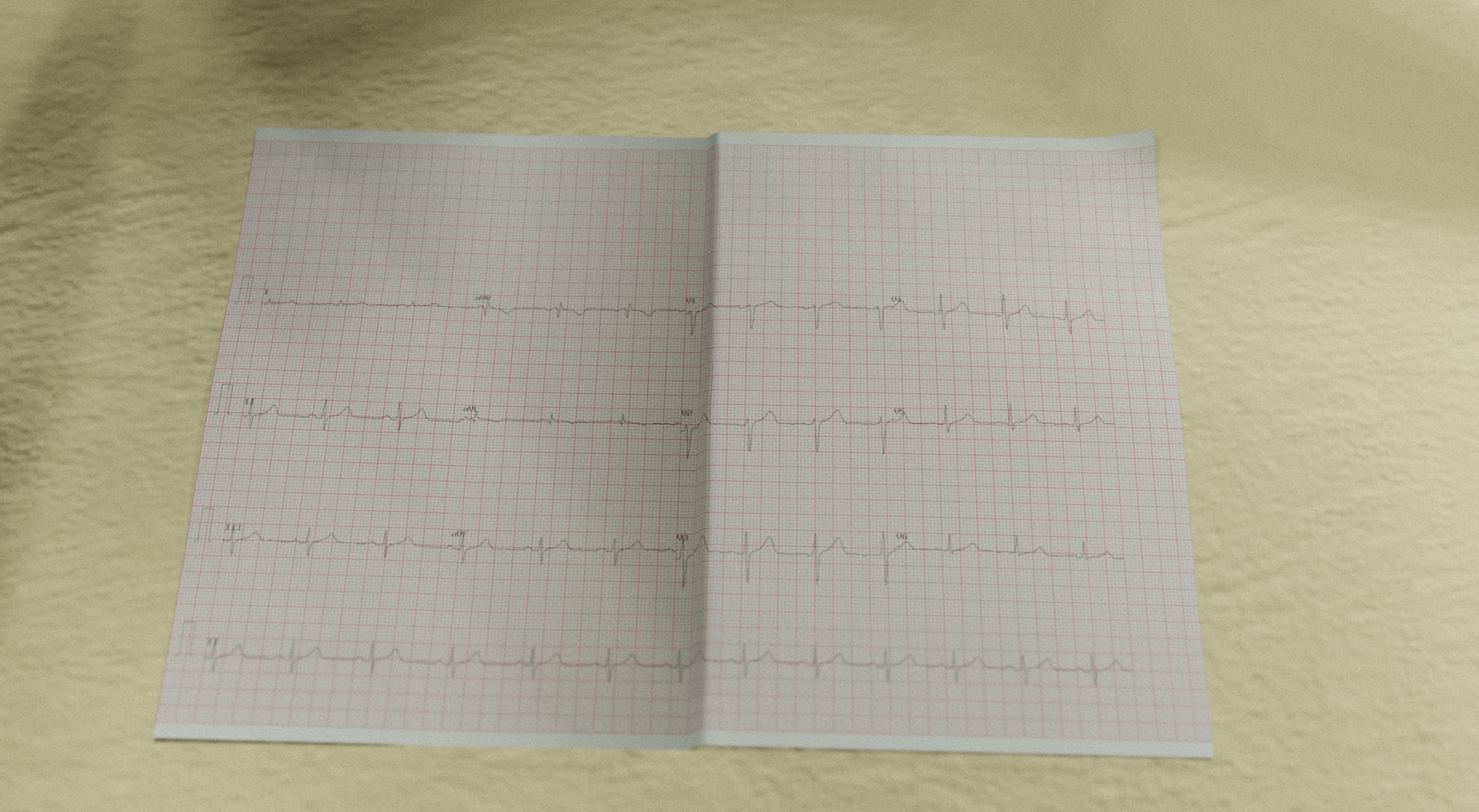}}
\caption{An example of a synthetically generated image from the GenECG dataset.}
\label{fig1}
\end{figure}

The ECG images used in this study were synthetically generated from the GenECG dataset, where lighting conditions often introduced contrast artifacts. To address this, we applied Contrast Limited Adaptive Histogram Equalization (CLAHE)~\cite{b0}, a technique to effectively normalize the image contrast. Subsequently, we utilized the segmentation-based tool Rembg~\cite{b1} to remove the background from the images, isolating the ECG paper. An example of an image from the GenECG dataset is shown in Fig.~\ref{fig1}.

After removing the background, a convex hull was fitted to the remaining ECG paper to accurately define its boundaries. The convex hull was simplified into a quadrilateral, and the corner points of this quadrilateral were used to compute a perspective transformation. 

The perspective transform maps the points on the original image $(x, y)$ to points on the warped image $(x', y')$ using a $3 \times 3$ homography matrix $\mathbf{H}$, as defined in Eq.~\ref{eq:perspective_transform}:

\begin{equation}
\begin{bmatrix}
x' \\
y' \\
w
\end{bmatrix}
=
\mathbf{H}
\begin{bmatrix}
x \\
y \\
1
\end{bmatrix},
\label{eq:perspective_transform}
\end{equation}

where $\mathbf{H}$ is the homography matrix:

\begin{equation}
\mathbf{H} =
\begin{bmatrix}
h_{11} & h_{12} & h_{13} \\
h_{21} & h_{22} & h_{23} \\
h_{31} & h_{32} & 1
\end{bmatrix}.
\label{eq:homography_matrix}
\end{equation}

The transformed coordinates $(x', y')$ are obtained by normalizing the homogeneous coordinates, as shown in Eq.~\ref{eq:normalized_coordinates}:

\begin{equation}
x' = \frac{h_{11}x + h_{12}y + h_{13}}{h_{31}x + h_{32}y + 1}, \quad
y' = \frac{h_{21}x + h_{22}y + h_{23}}{h_{31}x + h_{32}y + 1}.
\label{eq:normalized_coordinates}
\end{equation}

\begin{figure}[htbp]
\centerline{\includegraphics[width=8cm]{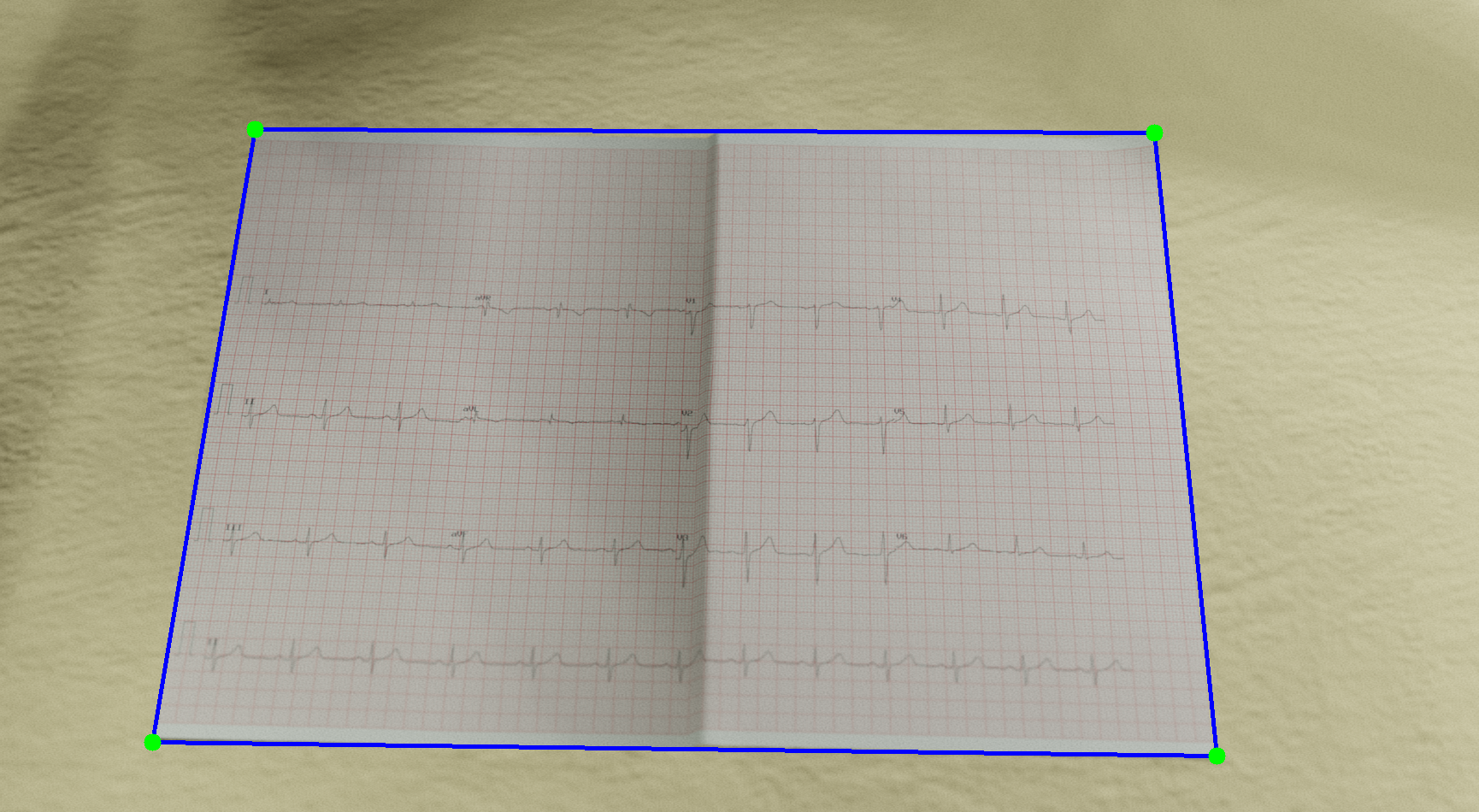}}
\caption{Simplified convex hull and corner points on the example image.}
\label{fig1_a}
\end{figure}

\begin{figure}[htbp]
\centerline{\includegraphics[width=8cm]{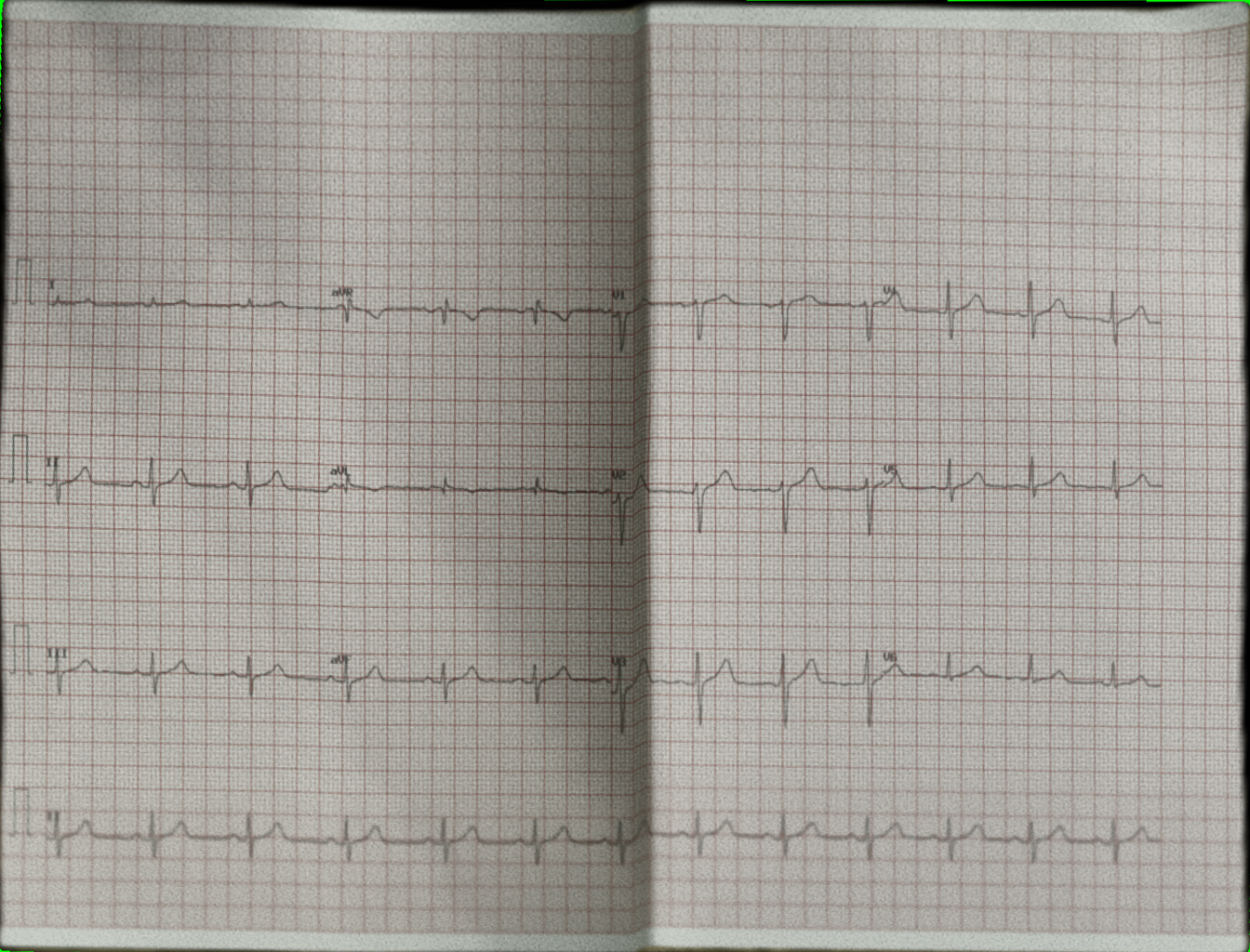}}
\caption{The processed example image after applying preprocessing steps, including background removal and perspective transformation.}
\label{fig2}
\end{figure}

This process ensures that the output image contains only the ECG paper with minimal residual rotation. The simplified convex hull and its corner points are illustrated in Fig.~\ref{fig1_a}, while the final warped image after preprocessing is shown in Fig.~\ref{fig2}.

\subsection{Model Development and Training}

For model training, we adopted a curriculum learning-based approach. Initially, a model was trained to classify cardiovascular diseases (CVDs) using segmentation masks, a relatively simpler task. The weights from this initial model were then transferred to a new model designed to operate on grayscale ECG images. The following subsections describe the details of this methodology.

\subsubsection{Training a Disease Classification Model from Segmentation Masks}

\begin{figure}[htbp]
\centerline{\includegraphics[width=8cm]{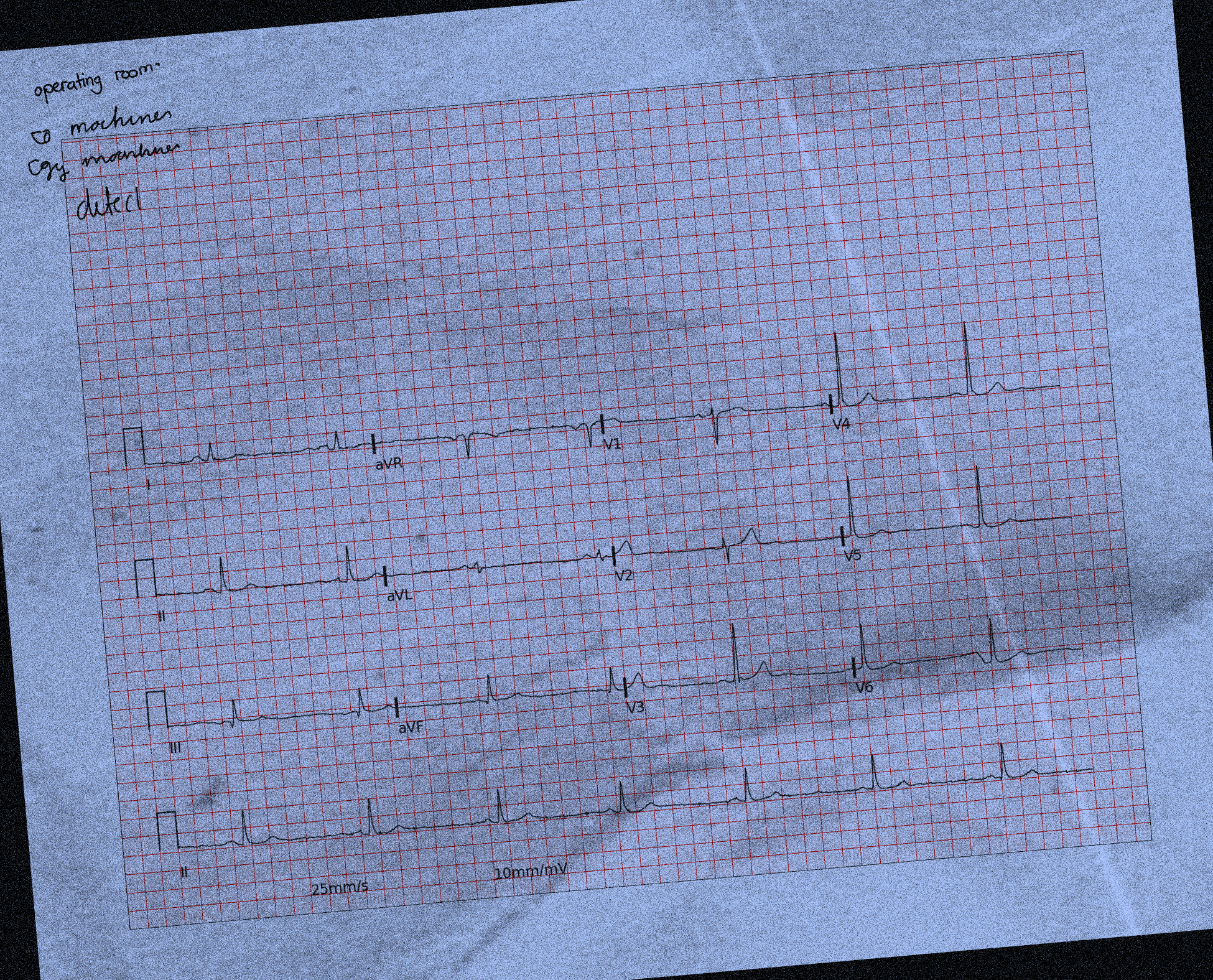}}
\caption{An example image generated by the ECG-image-kit using an ECG signal from the PTB-XL dataset.}
\label{fig_img_kit}
\end{figure}

As the first step, an EfficientNetV2-S~\cite{b8} model was trained on segmentation masks for disease classification. Since the GenECG dataset lacks ground truth segmentation masks, a U-Net~\cite{b9} model was trained to generate them. The training utilized a synthetic dataset created from PTB-XL samples with the ECG-image-kit~\cite{b10}, a tool capable of producing realistic ECG images with artifacts such as wrinkles, creases, rotations, noise, and handwritten text to simulate real-world conditions. A total of 2,855 samples were selected, with 90\% allocated for training and 10\% for testing. An example image generated by the ECG-image-kit is shown in Fig.~\ref{fig_img_kit}.

To address domain differences between synthetic and GenECG data, a self-supervised learning strategy was employed. The trained U-Net model was used to label 29 samples from the GenECG dataset, and these labeled samples were added to the training set. The segmentation model was trained using Focal Tversky Loss (FTL)~\cite{b12}, defined as:

\begin{equation}
    \text{FTL}_c = \sum_{c} (1 - TI_c)^{1/\gamma}
\end{equation}

where the Tversky index for class $c$ ($TI_c$) is given by:

\begin{equation}
    TI_c = \frac{\sum_{i=1}^{N} p_i g_i + \epsilon}{\sum_{i=1}^{N} p_i g_i + \alpha \sum_{i=1}^{N} p_i \overline{g_i} + \beta \sum_{i=1}^{N} \overline{p_i} g_i + \epsilon}.
\end{equation}

Here, \(p_i\) and \(g_i\) represent the predicted and ground truth values for pixel \(i\), respectively. The constants were set as \(\alpha = 1\), \(\beta = 10\), \(\gamma = 0.75\), and \(\epsilon = 10^{-6}\). The high \(\beta\) value emphasized the reduction of false negatives, ensuring that critical ECG features were preserved. To mitigate the resulting false positives, sliding window filtering and connected component analysis were employed.

The sliding window filtering algorithm processed the binary mask by dividing it into non-overlapping horizontal windows of fixed height (e.g., 10 pixels). Each window was examined to determine if it contained any signal (i.e., at least one non-zero pixel). Windows with signals were retained in the filtered mask, while those without were discarded. This approach ensured that only regions of the mask with relevant features were preserved while removing noise and false-positive regions. 

\begin{figure}[htbp]
\centerline{\includegraphics[width=8cm]{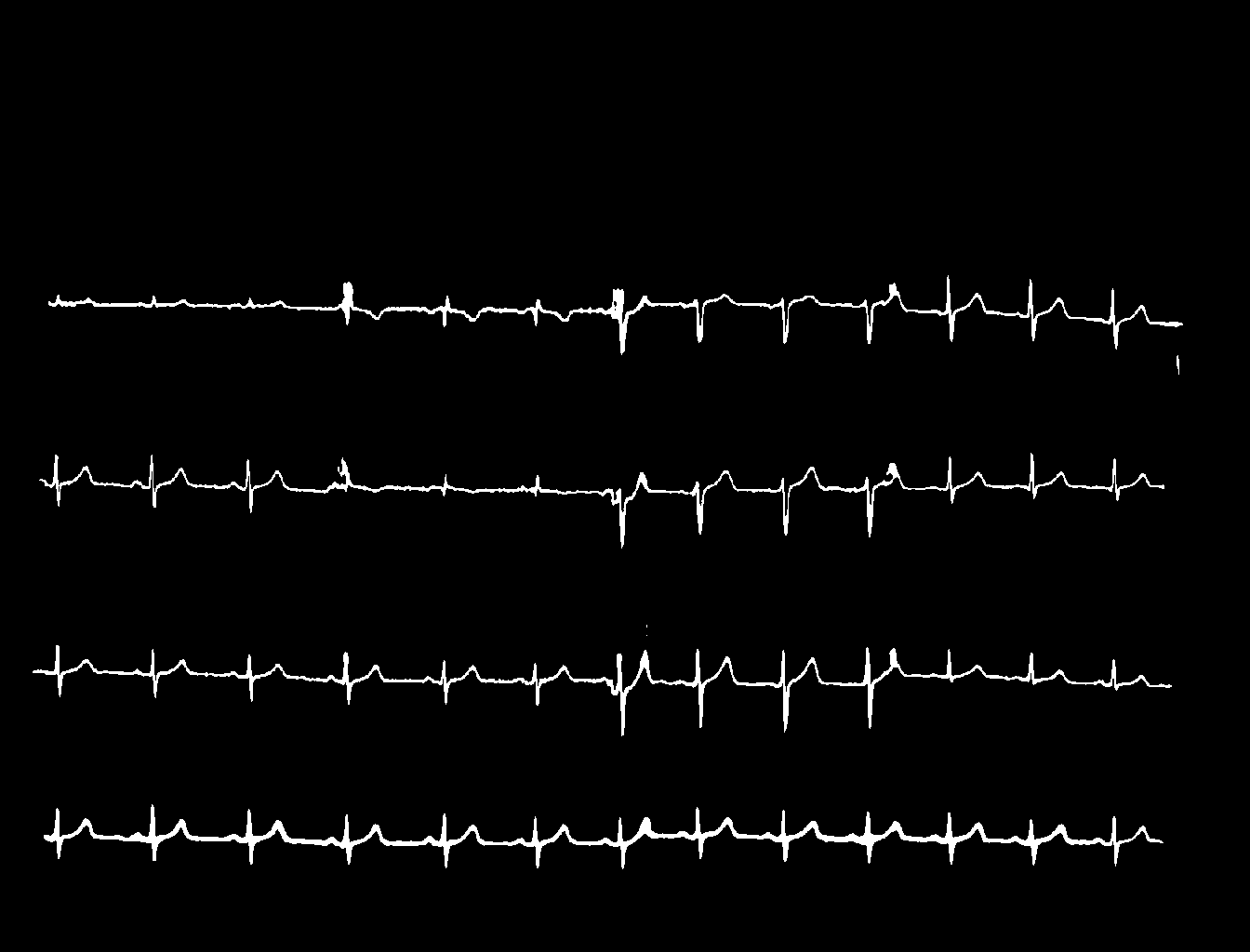}}
\caption{Segmentation mask generated from the preprocessed image.}
\label{fig_seg_mask}
\end{figure}

For connected component analysis, the Spaghetti Labeling algorithm~\cite{ConnectedComp} with 8-way connectivity was applied. This step further refined the segmentation by grouping connected regions of detected signal. The final segmentation mask for an example image, after applying these post-processing steps, is presented in Fig.~\ref{fig_seg_mask}.

Using these refined segmentation masks, an EfficientNetV2-S classifier was trained with binary cross-entropy with logits loss, defined as:

\begin{equation}
    \text{BCE}(y, \hat{y}) = -\frac{1}{N} \sum_{i=1}^{N} \left[ y_i \log \sigma(\hat{y}_i) + (1 - y_i) \log (1 - \sigma(\hat{y}_i)) \right],
\end{equation}

where $y_i$ is the ground truth label, $\hat{y}_i$ represents the predicted logits, $\sigma$ is the sigmoid function, and $N$ is the total number of samples. The classifier was trained using a Cosine Annealing Learning Rate Scheduler~\cite{b13} with an initial learning rate of 0.001 over 100 epochs. Random pixel dropout and rotation augmentations were applied to mitigate overfitting.

\subsubsection{Adapting the Model for Preprocessed ECG Images}

The weights of the initial model were transferred to a new model with the same architecture to facilitate its adaptation to grayscale ECG images. The warped ECG images were converted to grayscale, and their colors were inverted to resemble segmentation masks, improving convergence. The new classifier model was trained on these grayscale images using the same binary cross-entropy with logits loss.

\begin{figure}[htbp]
\centerline{\includegraphics[width=8cm]{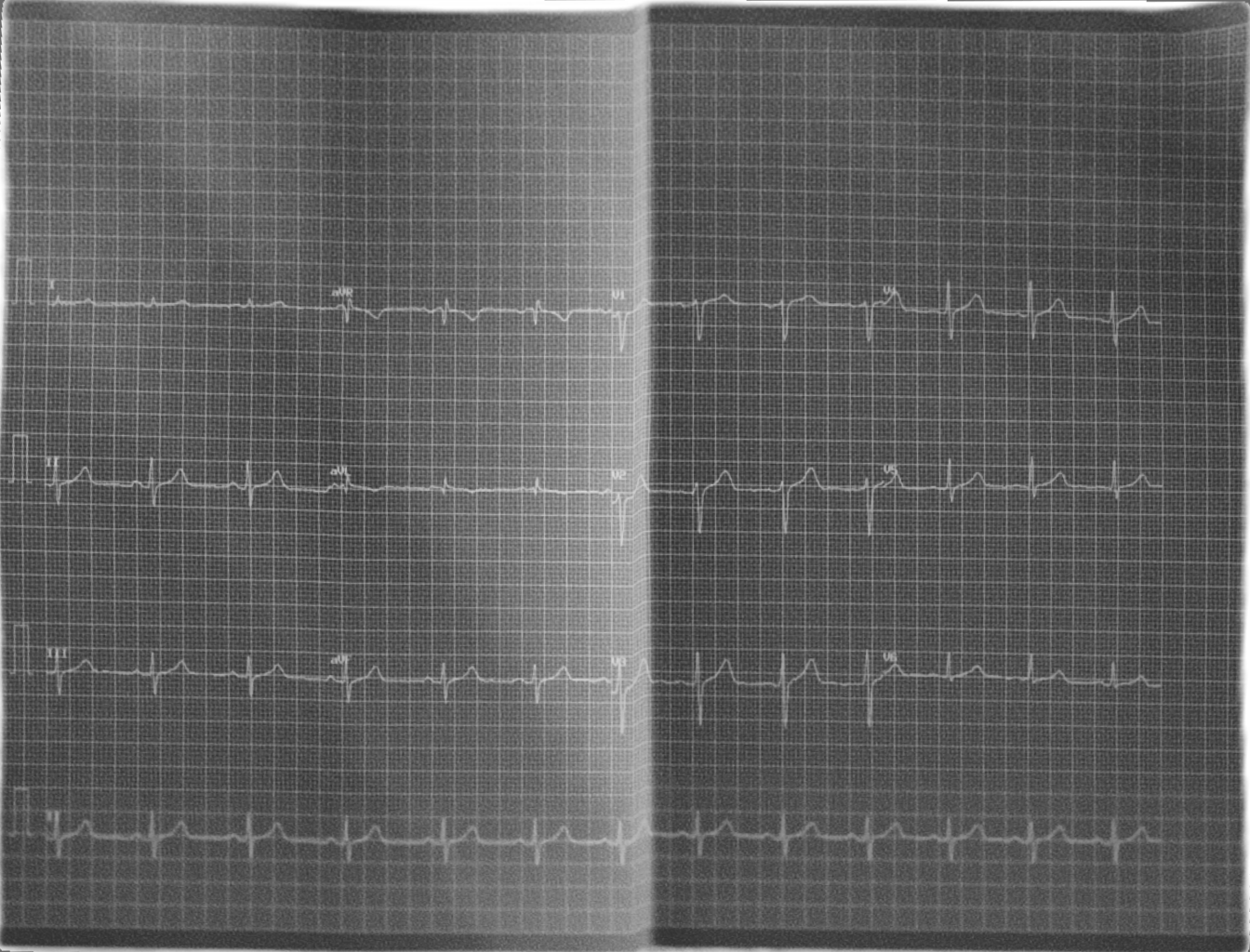}}
\caption{Example image after preprocessing and grayscale conversion.}
\label{fig_inverse}
\end{figure}

With the successful training of the new model, the segmentation step was eliminated, thereby simplifying the process. An example grayscale ECG image, obtained by transforming the image in Fig.~\ref{fig1}, is shown in Fig.~\ref{fig_inverse}.

\subsection{Ensemble Learning}
To enhance robustness, we employed an ensemble learning approach by training three versions of the final model, each with different hyperparameters and augmentation settings, as detailed in Table~\ref{tab:hyperparams}. This approach ensured variability in the learned representations, thereby improving generalization. The outputs of the individual models were combined by averaging their predictions, resulting in enhanced overall performance and reliability.

This two-step, ensemble-based methodology provided a robust and accurate diagnosis directly from ECG images, effectively addressing the challenges posed by real-world clinical data.

\begin{table}[h]
\centering
\caption{Hyperparameter and Augmentation Settings}
\label{tab:hyperparams}
\begin{tabular}{|c|c|c|c|}
\hline
\textbf{Model} & \textbf{Batch} & \textbf{Pixel Drop} & \textbf{Rotation} \\ 
               & \textbf{Size}  & \textbf{(Prob., Pixel-wise Prob.)}    & \textbf{(Limit, Prob.)} \\ 
\hline
Model 1 & 5  & 0.8, 0.01 & 10, 0.5 \\ 
\hline
Model 2 & 16 & 0.8, 0.01 & 30, 0.5 \\ 
\hline
Model 3 & 16 & 1.0, 0.01 & 30, 0.5 \\ 
\hline
\end{tabular}
\end{table}

\section{Experimental Results}

The experimental results, summarized in Table~\ref{tab:results}, were obtained on the BHF Data Science Centre ECG Challenge test set. Since the challenge evaluation metric is the Area Under the Curve (AUROC), the results were computed using the logits rather than the binarized outputs. The challenge test set is divided into two parts: public and private. Consequently, two separate scores are reported for each model, corresponding to the public and private portions of the test set.

The table includes the performance of three individual models and their ensemble. The ensemble, which combines the outputs of all three models by averaging their logits, achieved the highest scores, with an AUROC score of 0.93786 on the private portion of the held-out test set and 0.94715 on the public portion. These results demonstrate that the ensemble approach significantly improves the model's robustness and overall performance compared to individual models.
\begin{table}[h]
\centering
\caption{Experimental Results on the BHF Challenge Test Set}
\label{tab:results}
\begin{tabular}{|c|c|c|}
\hline
\textbf{Model}                   & \textbf{Private AUROC} & \textbf{Public AUROC} \\ 
\hline
Model 1                          & 0.92789                & 0.93731               \\ 
\hline
Model 2                          & 0.93199                & 0.93719               \\ 
\hline
Model 3                          & 0.92948                & 0.93579               \\ 
\hline
Ensemble of three models         & \textbf{0.93786}       & \textbf{0.94715}      \\ 
\hline
\end{tabular}
\end{table}

The validation set results are presented in Tables~\ref{tab:AUROC_results} and~\ref{tab:f1_results}. Table~\ref{tab:AUROC_results} reports the overall AUROC and per-class AUROC for each model and their ensemble. Among the individual models, Model 2 achieved the highest AUROC of 0.9530, while the ensemble model outperformed all, with an AUROC of 0.9534 and per-class AUROC ranging from 0.9302 to 0.9947.

Table~\ref{tab:f1_results} presents the F1 scores and per-class F1 scores, calculated by binarizing the logits using pre-determined per-class thresholds. The ensemble model achieved the best overall F1 score of 0.7801, with per-class F1 scores varying from 0.6343 to 0.8768. These results demonstrate the effectiveness of the ensemble approach in improving both AUROC and F1 metrics, providing a more robust classification system compared to individual models.

\begin{table}[h]
\centering
\caption{Validation Set Results: AUROC and Per-Class AUROC}
\label{tab:AUROC_results}
\begin{tabular}{|c|c|c|}
\hline
\textbf{Model} & \textbf{AUROC} & \textbf{Per-Class AUROC} \\ 
               &                & \textit{[STTC, HYP, MI, CD, AF]} \\ 
\hline
Model 1 & 0.9381 & [0.9129, 0.9157, 0.9304, 0.9396, 0.9919] \\ 
\hline
Model 2 & 0.9530 & [\textbf{0.9359}, 0.9263, 0.9516, 0.9572, 0.9938] \\ 
\hline
Model 3 & 0.9506 & [0.9327, 0.9250, 0.9485, 0.9529, 0.9939] \\ 
\hline
Ensemble & \textbf{0.9534} & [0.9320, \textbf{0.9302}, \textbf{0.9523}, \textbf{0.9575}, \textbf{0.9947}] \\ 
\hline
\end{tabular}
\end{table}

\begin{table}[h]
\centering
\caption{Validation Set Results: F1 Scores and Per-Class F1 Scores}
\label{tab:f1_results}
\begin{tabular}{|c|c|c|}
\hline
\textbf{Model} & \textbf{F1 Score} & \textbf{Per-Class F1 Score} \\ 
               &                & \textit{[STTC, HYP, MI, CD, AF]} \\ 
\hline
Model 1 & 0.7486 & [0.7255, 0.6197, 0.7796, 0.7802, 0.8381] \\ 
\hline
Model 2 & 0.7797 & [\textbf{0.7680}, 0.6313, 0.8197, 0.8025, \textbf{0.8768}] \\ 
\hline
Model 3 & 0.7736 & [0.7601, 0.6328, 0.8091, 0.7915, 0.8744] \\ 
\hline
Ensemble & \textbf{0.7801} & [0.7597, \textbf{0.6343}, \textbf{0.8206}, \textbf{0.8091}, \textbf{0.8768}] \\ 
\hline
\end{tabular}
\end{table}

\begin{figure}[h]
\centering
\subfloat[]{\includegraphics[width=6.9cm]{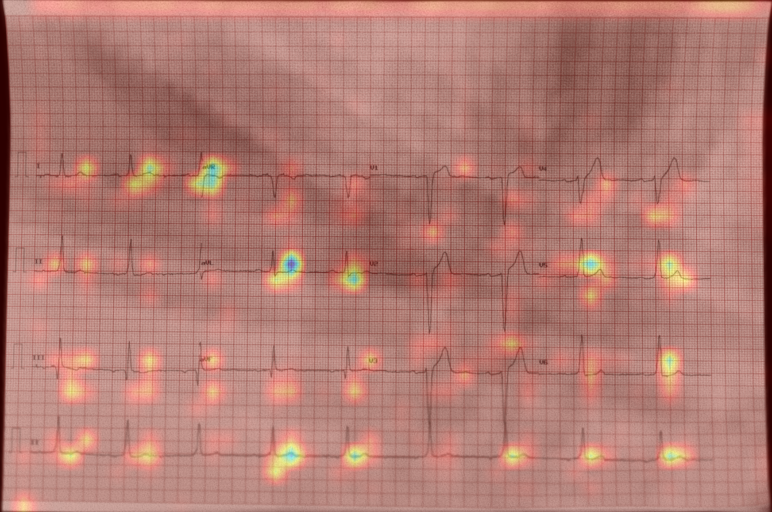}\label{gradcam}}\\ 
\subfloat[]{\includegraphics[width=6.9cm]{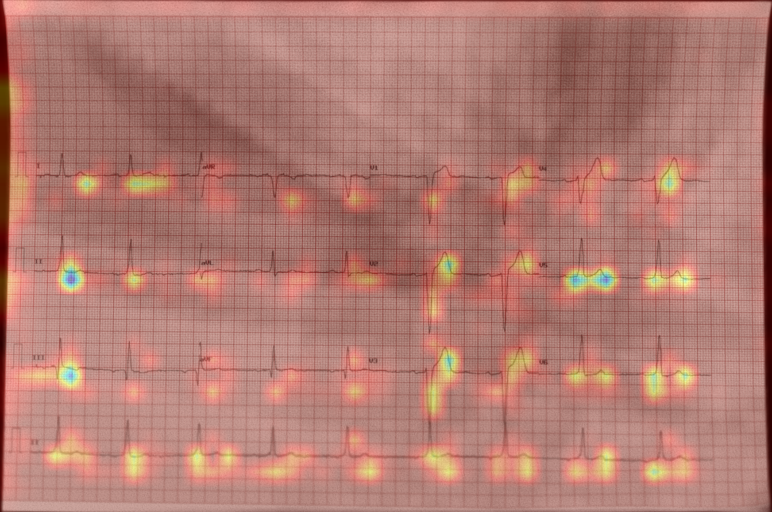}\label{gradcam_0_2}}
\\
\subfloat[]{\includegraphics[width=6.9cm]{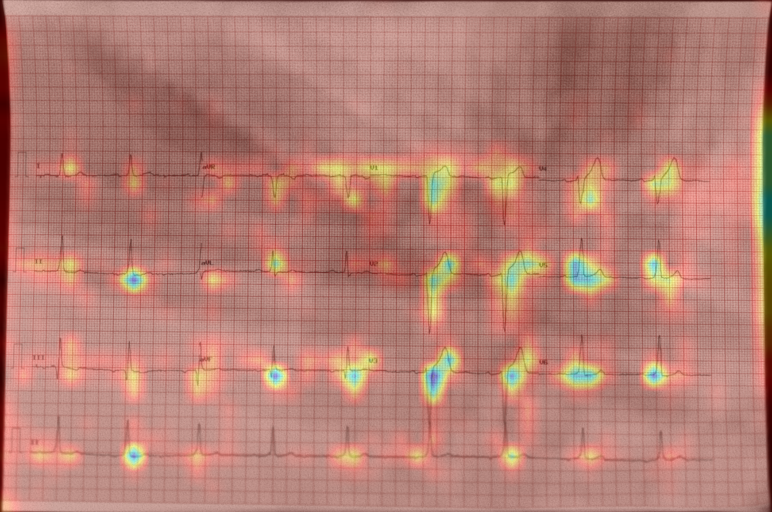}\label{gradcam_0_3}}
\caption{XGrad-CAM heatmaps for the STTC class, shown for \protect\subref{gradcam} Model 1, \protect\subref{gradcam_0_2} Model 2, and \protect\subref{gradcam_0_3} Model 3.}
\label{fig:Grad_class0}
\end{figure}

\begin{figure}[h]
\centering
\subfloat[]{\includegraphics[width=6.9cm]{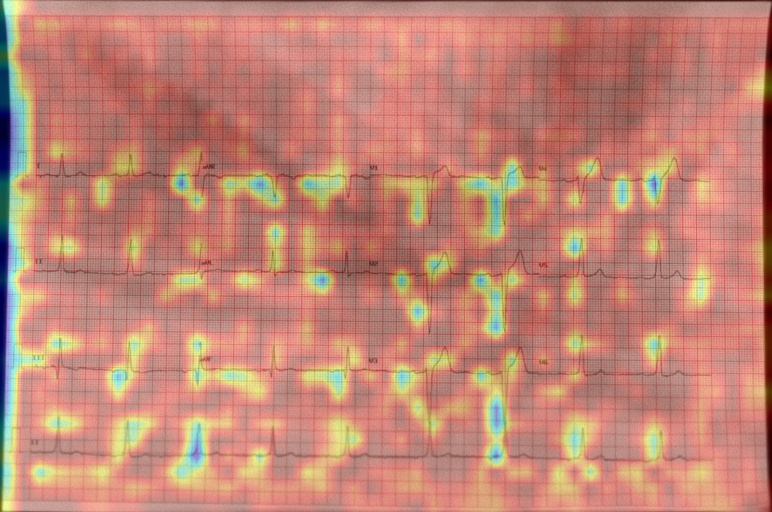}\label{gradcam_2}}
\\
\subfloat[]{\includegraphics[width=6.9cm]{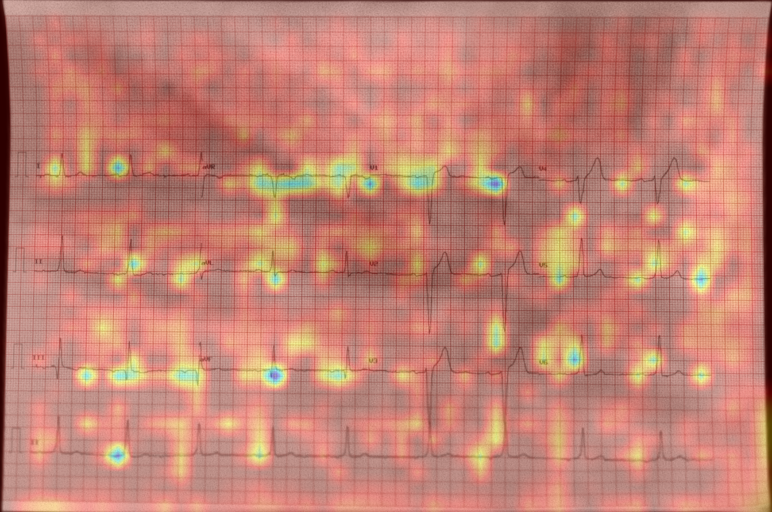}\label{gradcam_2_2}}
\\
\subfloat[]{\includegraphics[width=6.9cm]{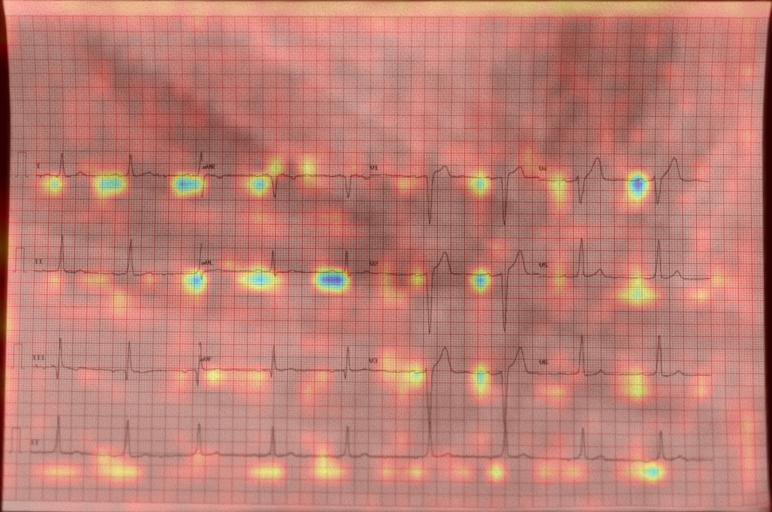}\label{gradcam_2_3}}
\caption{XGrad-CAM heatmaps for the MI class, shown for \protect\subref{gradcam_2} Model 1, \protect\subref{gradcam_2_2} Model 2, and \protect\subref{gradcam_2_3} Model 3.}
\label{fig:Grad_class}
\end{figure}

To analyze the regions of focus for each individual model, we applied the XGrad-CAM~\cite{b14} technique to the STTC and CD classes. The XGrad-CAM outputs for the STTC class are shown in Fig.~\ref{fig:Grad_class0}, while the outputs for the MI class are presented in Fig.~\ref{fig:Grad_class}. These figures reveal that each model focuses on different features within the printed ECGs, and the focus regions vary across classes. 

These findings highlight the effectiveness of the ensemble learning approach. As each model focuses on distinct features, their combined outputs average the information, leveraging complementary insights to improve overall performance. For instance, in Fig.~\ref{fig:Grad_class}\subref{gradcam_2}, it is evident that Model 1's focus is distracted by the artifact present on the left side of the image. However, this issue is not observed in Fig.~\ref{fig:Grad_class}\subref{gradcam_2_2} and Fig.~\ref{fig:Grad_class}\subref{gradcam_2_3}, which demonstrates how ensemble learning mitigates such individual model weaknesses and enhances overall robustness.

\section{Conclusion}
This study introduced a novel method for direct cardiovascular disease (CVD) diagnosis from ECG images, eliminating the need for digitization. The proposed approach employs a two-step framework: pre-training on segmentation masks followed by fine-tuning on grayscale, inverted ECG images. By leveraging ensemble learning, we improved robustness and achieved high performance on the validation set, with an AUC of 0.9534 and an F1 score of 0.7801. On the held-out test set, the ensemble model attained an AUROC score of 0.93786 on the private part and 0.94715 on the public part. The gains from the ensemble learning approach were more pronounced on the test set, demonstrating increased robustness and better generalization to unseen data.

Our analysis using XGrad-CAM revealed that individual models focus on distinct features within ECG images, with variations across classes. This diversity in focus underscores the strength of the ensemble learning approach, as combining these complementary insights mitigates weaknesses in individual models and enhances overall reliability.

While the proposed method performs well overall, it exhibits relatively lower performance on the HYP class. However, this is less critical in practice, as HYP generally does not require urgent intervention unless observed with other classes. In contrast, STTC, MI, and AF classes—associated with conditions requiring rapid diagnosis—are prioritized and performed better, aligning with the goal of this work to provide timely and efficient diagnostics without reliance on more complex tools.

In future work, we aim to focus on improving the explainability of the model's decisions. Enhancing explainability will provide clinicians with greater insight into the model's reasoning, fostering trust and facilitating integration into clinical workflows. This could involve developing advanced interpretability methods or visualizations that highlight diagnostically significant features in ECG images, particularly for critical conditions like MI and AF.

\end{document}